\documentclass[sigconf, nonacm]{acmart}

\newcommand\vldbdoi{XX.XX/XXX.XX}

\newcommand\vldbavailabilityurl{URL_TO_YOUR_ARTIFACTS}

\begin{document}
\title{TF-Engram: A Train-Free Engram with SSD-Backed Memory for Large Language Models}

\author{Yutang MA}
\affiliation{%
  \institution{The Chinese University of Hong Kong}
}
\email{ytma2@cse.cuhk.edu.hk}

\author{Kecheng HUANG}
\affiliation{%
  \institution{Beijing Institute of Technology, Zhuhai}
}
\email{huangkecheng@bitzh.edu.cn}

\author{Xikun JIANG}
\affiliation{%
  \institution{The Chinese University of Hong Kong}
}
\email{xkjiang24@cse.cuhk.edu.hk}

\author{Zili SHAO}
\affiliation{%
  \institution{The Chinese University of Hong Kong}
}
\email{shao@cse.cuhk.edu.hk}

\begin{abstract}
Large Language Models (LLMs) store factual knowledge and domain-specific patterns implicitly in dense Transformer parameters, making knowledge expansion costly through pretraining, fine-tuning, retrieval augmentation, or longer contexts. Engram-style memory offers a compact hidden-state injection pathway, but existing GPU-resident designs often rely on hash-based compression, causing unrelated phrases to collide in shared slots and weakening phrase-level semantic fidelity. We present TF-Engram, a train-free Engram system that constructs phrase-specific semantic memory offline from external corpora, stores large memory tables across a GPU--DRAM--SSD hierarchy, and uses Early-Exit Guided Predictive Prefetching to hide external-memory latency during autoregressive decoding. On Qwen3-0.6B, TF-Engram improves the average downstream score from 57.6 to 59.4, outperforming both the frozen backbone and a parameter-matched LoRA baseline. System evaluation shows that large TF-Engram tables can be built with moderate offline cost, SSD-backed storage substantially reduces GPU memory demand, and predictive prefetching recovers much of the throughput loss caused by external memory access. These results demonstrate that static phrase memory can be integrated into LLM inference as a scalable, train-free, and low-overhead system component.
\end{abstract}

\maketitle



\section{Introduction}

Large Language Models (LLMs) have become the dominant architecture for natural language understanding, reasoning, code generation, and knowledge-intensive question answering~\cite{vaswani2017attention,devlin2019bert,brown2020language,chowdhery2022palm,touvron2023llama,touvron2023llama2}. Their success is largely driven by scaling dense Transformer parameters and training data~\cite{kaplan2020scaling,hoffmann2022training}. However, this design also tightly couples knowledge capacity with model size: factual knowledge, domain-specific terminology, and frequently occurring semantic patterns are implicitly stored inside billions of neural parameters. Improving such knowledge often requires more pretraining, supervised fine-tuning, retrieval augmentation, or longer context windows~\cite{brown2020language,lewis2020rag,guu2020realm,borgeaud2022retro}. These solutions increase training cost, inference latency, memory consumption, or deployment complexity. As LLMs are increasingly used in domain-specific and knowledge-intensive applications, it becomes important to decouple static knowledge storage from dense Transformer computation.

Recent work on memory-augmented LLMs explores this decoupling from several directions. Retrieval-Augmented Generation (RAG) systems maintain an external document corpus and retrieve relevant passages during inference, usually appending the retrieved text to the prompt~\cite{lewis2020rag,guu2020realm,borgeaud2022retro,izacard2022atlas}. Long-context and KV-cache based methods extend the amount of information accessible to the model by keeping more historical tokens or intermediate states~\cite{dao2022flashattention,dao2024flashattention2,kwon2023vllm}. Adapter-based and parameter-efficient memory methods inject additional trainable modules into the model, allowing new knowledge to be stored in a smaller number of parameters than full fine-tuning~\cite{houlsby2019adapters,li2021prefixtuning,lester2021prompttuning,hu2021lora}. Another line of work, represented by Engram-style memory, introduces a lightweight side memory pathway that stores token-level or phrase-level patterns and injects retrieved memory signals into the hidden states of the backbone model~\cite{memorygrafting}. These categories differ substantially in memory granularity, training cost, inference overhead, and system scalability.

Although RAG, long-context, and adapter-based methods are effective in many scenarios, they are not ideal for fine-grained static phrase memory. RAG retrieves coarse-grained text chunks rather than compact phrase-level memory entries, and the retrieved passages increase context length and attention cost~\cite{lewis2020rag,borgeaud2022retro,izacard2022atlas}. Long-context methods allow the model to attend to more information, but they still require the Transformer to process additional tokens, even when the useful knowledge corresponds to short and stable phrases~\cite{dao2022flashattention,dao2024flashattention2}. Adapter-based memory methods reduce the cost of full fine-tuning, but they still require training and are usually tied to a specific backbone model~\cite{houlsby2019adapters,li2021prefixtuning,lester2021prompttuning,hu2021lora}. In contrast, Engram-style memory is a more suitable starting point for a system-oriented design. It provides a compact side pathway, operates closer to token or phrase granularity, and can be integrated into LLM inference without restructuring the main Transformer computation~\cite{memorygrafting}. Therefore, this paper focuses on extending Engram-style memory into a scalable external memory system for LLMs.

Despite its promise, Engram-style memory is not yet a practical large-scale static memory system. Existing designs are mainly optimized as compact GPU-resident neural memory modules. To fit a large number of token patterns into limited accelerator memory, they often rely on hash-based compression, where many token sequences are mapped into a fixed-size table. This design reduces memory footprint, but it introduces a tension between memory coverage and memory fidelity. To make Engram suitable for large static phrase knowledge, three critical issues must be addressed: how to construct phrase-specific memory without additional training, how to scale memory capacity without relying on ambiguous hash-collision slots, and how to access external memory without stalling autoregressive decoding.

The first critical issue is that memory construction remains training-dependent and model-specific. Conventional Engram designs typically learn memory entries, projection layers, gates, or auxiliary parameters from data, either jointly with the backbone model or through additional post-training~\cite{memorygrafting,houlsby2019adapters,hu2021lora}. This makes the memory difficult to deploy across different LLMs or adapt to new domains without extra optimization. Moreover, when memory entries are constructed at token or hash-slot granularity, the resulting values do not explicitly preserve phrase-level semantics. A scalable Engram system should therefore be able to construct phrase-level memory offline, preserve semantic information, and avoid retraining the backbone model.

The second critical issue is the trade-off between capacity and semantic fidelity. Existing Engram-style memories can remain compact by mapping many token patterns into a fixed-size hash table. However, when the candidate phrase space is much larger than the table, unrelated phrases inevitably collide in the same memory slots. A single slot may then receive signals from multiple unrelated phrases, causing its memory value to become an ambiguous mixture rather than a clean phrase-specific representation. This weakens the original goal of static phrase memory: a phrase such as \emph{New York City}, \emph{quantum field theory}, or \emph{BGP route reflector} should ideally retrieve a stable memory signal tied to that phrase, not a value shared with many unrelated patterns. Therefore, simply compressing more phrases into a small GPU table increases apparent coverage but can reduce the usefulness of the retrieved memory. Scaling Engram memory requires preserving precise phrase entries and moving the capacity burden into a storage hierarchy.

The third critical issue is access latency after precise memory is externalized. Moving memory entries from GPU memory to host DRAM or NVMe SSD can greatly expand capacity and reduce hash collisions, but naive external lookup can break the latency requirements of autoregressive decoding. During generation, each token depends on previous outputs, and the model has limited opportunity to wait for random storage accesses. Prior LLM serving and offloading systems show that memory hierarchy, scheduling, and data movement are central to efficient inference~\cite{yu2022orca,kwon2023vllm,sheng2023flexgen,aminabadi2022deepspeed}. A straightforward prefetch design may use the final next-token distribution produced by the original LLM head. However, this signal becomes available only after the full forward pass completes, leaving little remaining computation to overlap with SSD or CPU--GPU transfer. Therefore, SSD-backed Engram memory requires an earlier prediction mechanism that can anticipate future memory requests before the final layers finish.

To address these issues, we propose \textbf{TF-Engram}, a train-free Engram with SSD-backed memory for large language models. The key idea is to shift Engram from a small hash-compressed learned table into a storage-managed external memory system. Instead of forcing a large phrase space into ambiguous shared slots, TF-Engram constructs phrase-specific semantic memory offline, places it across a GPU--DRAM--SSD hierarchy, and uses early-layer prediction signals to prefetch future memory entries before they are needed.

To address the first issue, TF-Engram constructs semantic phrase memory entirely offline. It first mines candidate phrases from large-scale corpora, such as general web text, encyclopedic text, and domain-specific scientific corpora. These phrases are then tokenized using the target LLM tokenizer, encoded using an external semantic encoder, and stored as static key-value memory entries. Unlike conventional Engram memories that learn memory parameters from model training, TF-Engram separates memory construction from backbone optimization. This design allows the memory table to be built once, reused across inference runs, and extended with new corpora without retraining the LLM itself.

To address the second issue, TF-Engram organizes memory as an SSD-backed hierarchy. Frequently accessed entries are kept in GPU or host-memory caches for low-latency lookup, while cold entries are offloaded to commodity NVMe SSDs. This hierarchy allows memory capacity to scale independently from accelerator memory. Instead of using hash compression as the main mechanism for fitting memory into GPU, TF-Engram preserves phrase-specific logical entries and relies on hierarchical storage to absorb the capacity demand. This design makes it possible to support much larger phrase tables while reserving scarce GPU memory for model weights, activations, KV cache, and the hottest memory entries~\cite{kwon2023vllm,sheng2023flexgen,rajbhandari2021zeroinfinity}.

To address the third issue, TF-Engram introduces Early-Exit Guided Predictive Prefetching. Rather than waiting for the final language modeling head to produce the next-token distribution, TF-Engram attaches an auxiliary prefetch head to an intermediate layer near the end of the LLM, such as layer $L-r$. This early-exit head produces an approximate next-token distribution before the full forward pass finishes. TF-Engram expands the predicted top-$K$ tokens into likely future phrase accesses and asynchronously prefetches their memory entries from SSD or host DRAM while the remaining Transformer layers continue computation. By using the remaining layers as a latency-hiding window, TF-Engram reduces blocking external memory accesses and makes SSD-backed phrase memory practical for autoregressive inference.

Our evaluation on Qwen3-0.6B shows that TF-Engram improves end-to-end model quality while keeping the backbone frozen. Across ten downstream benchmarks, TF-Engram increases the average score from 57.6 to 59.4, outperforming both the original backbone and a parameter-matched LoRA baseline. The system analysis further shows that large TF-Engram tables can be constructed with moderate offline cost, while the SSD-backed hierarchy substantially reduces GPU memory demand compared with GPU-resident memory. With early-exit guided prefetching, TF-Engram recovers much of the throughput loss caused by external memory access and keeps decoding latency overhead modest. These results indicate that TF-Engram provides a practical path toward train-free, scalable, and low-overhead memory augmentation for LLM inference.

\section{Background}

\subsection{Memory-Augmented LLMs}

Large Language Models (LLMs) are typically implemented as decoder-only Transformers~\cite{vaswani2017attention,brown2020language,touvron2023llama,touvron2023llama2}. Given an input sequence $x = {x_1, x_2, \ldots, x_t}$, the model first maps token IDs into continuous embeddings and then processes them through a stack of Transformer layers. At each layer, the model refines the hidden representation of every token by applying self-attention and feed-forward transformations~\cite{vaswani2017attention}. After the final layer, the hidden state at position $t$ is projected by the language modeling head to produce a probability distribution over the next token:
\begin{equation}
p(x_{t+1} \mid x_{\leq t}) = \mathrm{softmax}(W_{\mathrm{lm}} h_t^{L}),
\end{equation}
where $h_t^{L}$ denotes the final-layer hidden state and $L$ is the number of Transformer layers.

During autoregressive decoding, this process is repeated token by token~\cite{brown2020language,touvron2023llama2}. Since each generated token depends on previously generated tokens, LLM inference is inherently sequential at the decoding stage. Modern inference systems reduce redundant computation by caching previous key-value states, but each new token still requires a forward pass through the model~\cite{kwon2023vllm,yu2022orca}. Therefore, any additional operation placed on the per-token critical path can directly affect latency and throughput. This latency-sensitive property is especially important when integrating external memory, because memory lookup, data movement, and memory injection may all become part of the decoding pipeline.

A fundamental property of standard LLMs is that most knowledge is stored implicitly in dense parameters~\cite{brown2020language,kaplan2020scaling,hoffmann2022training}. Factual associations, entity knowledge, technical terminology, and recurring linguistic patterns are encoded through pretraining rather than stored in an explicit external structure. This design is powerful because it allows the model to generalize from large-scale data, but it also creates several system-level limitations. First, increasing knowledge capacity often requires increasing model size, which raises both training and inference cost~\cite{kaplan2020scaling,hoffmann2022training,chowdhery2022palm}. Second, updating knowledge is difficult because modifying a small set of facts may require fine-tuning or retraining~\cite{hu2021lora,houlsby2019adapters}. Third, static knowledge and dynamic computation are tightly coupled: even if a piece of knowledge is stable and repeatedly used, the model still retrieves it through dense Transformer computation.

Memory-augmented LLMs aim to address this limitation by introducing external or auxiliary memory~\cite{lewis2020rag,guu2020realm,borgeaud2022retro,khandelwal2020knnlm,memorygrafting}. Instead of relying solely on parametric knowledge, the model can access additional information through a memory pathway. Existing memory-augmented methods differ in what they store, how they access memory, and where the memory is integrated into the inference pipeline. From a system perspective, these methods can be roughly categorized into retrieval-based memory, context-based memory, parameter-based memory, and internal side-pathway memory.

Retrieval-Augmented Generation (RAG) is the most widely used form of external memory~\cite{lewis2020rag,guu2020realm,izacard2022atlas}. A RAG system maintains a corpus of documents or passages outside the model. Given an input query, the system computes a retrieval query, searches an external index, and appends the retrieved passages to the prompt. The LLM then consumes the original input together with the retrieved context. This approach is flexible because the corpus can be updated independently from the model parameters. It is also effective for knowledge-intensive tasks where relevant information can be found in documents~\cite{lewis2020rag,borgeaud2022retro,izacard2022atlas}. However, RAG usually operates at coarse granularity. The retrieved units are typically paragraphs or passages, not compact semantic memories. As a result, RAG increases prompt length and attention cost, even when the useful information is only a short phrase or entity. Moreover, retrieval and prompt construction are often outside the model's internal computation, making RAG less suitable for fine-grained token-level or phrase-level memory injection.

Context-based memory methods take a different approach. Long-context models, sliding-window attention, recurrent memory mechanisms, and KV-cache reuse techniques allow the model to access more historical information~\cite{dao2022flashattention,dao2024flashattention2,kwon2023vllm}. These methods are useful when relevant information already appears in the input context or previous conversation. However, they do not directly solve the problem of storing large amounts of reusable external knowledge. A long-context model still needs to process additional tokens through attention. This is expensive when the same static knowledge is repeatedly needed across many requests. KV-cache based methods preserve intermediate states for previously processed tokens, but the cached content is request-specific rather than a general-purpose memory table~\cite{kwon2023vllm}. Therefore, context-based memory is better suited for preserving dynamic conversational history than for storing large-scale static phrase knowledge.

Parameter-based memory methods store new information by adding or updating trainable parameters. Examples include adapters, LoRA-style modules, memory layers, and other parameter-efficient tuning mechanisms~\cite{houlsby2019adapters,li2021prefixtuning,lester2021prompttuning,hu2021lora}. Compared with full fine-tuning, these methods reduce the number of updated parameters and can be easier to deploy. They are useful when a model needs to adapt to a new domain or task. However, they still require optimization. The learned parameters are usually tied to a specific backbone architecture, hidden dimension, tokenizer, and training objective. If the backbone model changes, the memory module may need to be retrained or redesigned. Furthermore, parameter-based memory does not naturally provide explicit indexing or storage hierarchy: once knowledge is absorbed into parameters, it becomes difficult to selectively retrieve, update, evict, or offload.

Internal side-pathway memory methods introduce a more compact integration point~\cite{memorygrafting,khandelwal2020knnlm}. Instead of appending retrieved text to the prompt or modifying a large number of parameters, these methods add a memory pathway inside the model's inference process. The memory is usually represented as a key-value structure. During inference, the model forms a query from token history or hidden states, retrieves one or more memory values, and injects the retrieved signal into the hidden representation. This design has two important advantages. First, it avoids increasing the input context length. Second, it allows memory access to operate at a finer granularity than document retrieval. Engram-style memory belongs to this category~\cite{memorygrafting}.

Another useful property of decoder-only Transformers is that intermediate layers already contain partial predictive information~\cite{leviathan2023speculative,chen2023speculative}. Although the final next-token distribution is produced only after all $L$ layers finish, an intermediate hidden state $h_t^{\ell}$, where $\ell < L$, can also be projected to produce an approximate next-token distribution. Such an auxiliary prediction does not need to replace the final model output. Instead, it can be used for system-level scheduling decisions, such as prefetching likely future memory entries before the full forward pass completes. This property is important for latency-sensitive memory systems because it creates an opportunity to overlap memory transfer with the computation of the remaining Transformer layers.

These memory categories reflect different trade-offs. RAG provides flexible document-level access but adds retrieval and context-processing overhead~\cite{lewis2020rag,izacard2022atlas}. Long-context methods preserve more dynamic history but are expensive for static knowledge~\cite{dao2022flashattention,dao2024flashattention2}. Parameter-based methods can encode new information but require training and are model-specific~\cite{houlsby2019adapters,hu2021lora}. Side-pathway memory provides a compact mechanism for inference-time knowledge injection, making it attractive for system-level optimization~\cite{memorygrafting}. This paper focuses on Engram-style side-pathway memory because it is naturally aligned with phrase-level memory, does not require prompt expansion, and can potentially be implemented as an efficient external memory system.

\subsection{Engram-Style Memory}

Engram-style memory is designed to supplement the Transformer backbone with an auxiliary memory pathway\cite{cheng2026engram}. The core idea is to store reusable token-level or phrase-level patterns in a memory table and retrieve relevant memory signals during inference. Unlike RAG, which retrieves raw text and feeds it back into the model as additional context~\cite{lewis2020rag,borgeaud2022retro}, Engram-style memory retrieves hidden-space values and injects them directly into the model representation. This makes the memory more compact and avoids increasing the sequence length processed by attention.

A generic Engram-style memory can be abstracted as a key-value table:
\begin{equation}
\mathcal{M} = {(k_i, v_i)}_{i=1}^{N},
\end{equation}
where $k_i$ is a memory key and $v_i$ is the corresponding memory value. The key is used for lookup, while the value is used to modify the model's hidden state. Depending on the design, the key may be derived from token IDs, token $n$-grams, hashed token sequences, phrase identifiers, or hidden-state representations. The value is usually represented in the same or a compatible vector space as the LLM hidden states, so that it can be added to or fused with the model's internal representation~\cite{memorygrafting,khandelwal2020knnlm}.

During inference, the model computes a hidden state $h_t$ for the current position $t$. The memory module then derives a query $q_t$ from the current context. This query may depend on the recent token sequence, the current hidden state, or both. The memory lookup function returns a retrieved memory signal:
\begin{equation}
m_t = M(q_t),
\end{equation}
where $M(\cdot)$ denotes the memory lookup and aggregation function. If multiple memory entries are matched, the returned signal may be computed through weighted aggregation:
\begin{equation}
m_t = \sum_{i \in \mathcal{N}(q_t)} \alpha_i v_i,
\end{equation}
where $\mathcal{N}(q_t)$ is the set of retrieved entries and $\alpha_i$ is the retrieval weight for entry $i$.

The retrieved memory signal is then injected into the model. A common formulation is residual addition with a gate:
\begin{equation}
h'_t = h_t + g_t \cdot m_t,
\end{equation}
where $h'_t$ is the memory-augmented hidden state and $g_t$ controls how strongly the memory affects the model. The gate may be a scalar, a vector, or a learned function of the current hidden state and retrieved memory. This gating mechanism is important because not every memory retrieval is useful. When the memory signal is relevant, the gate can amplify it; when the memory is noisy or unnecessary, the gate can suppress it~\cite{memorygrafting}.

Engram-style memory has several properties that make it attractive for LLM systems. First, it is compact. Instead of storing full documents or long passages, it stores structured memory entries associated with token or phrase patterns. This allows memory to represent frequently reused knowledge units more efficiently. Second, it does not require prompt expansion. Retrieved memory values are injected into hidden states rather than appended to the input sequence, avoiding additional attention over retrieved text. Third, it can be integrated as a side pathway. The main Transformer computation remains largely unchanged, which makes the design compatible with existing decoder-only LLM architectures~\cite{vaswani2017attention,brown2020language,touvron2023llama2}.

A practical Engram-style system must also control memory size. The number of possible token $n$-grams or phrase patterns grows rapidly with corpus size. Directly assigning an independent vector to every possible pattern would quickly exceed GPU memory. To keep the memory table compact, existing designs often use hash-based compression. A hash function maps many token sequences into a fixed-size table:
\begin{equation}
b = \mathrm{Hash}(\tau) \bmod B,
\end{equation}
where $\tau$ is a token sequence, $B$ is the number of physical memory slots, and $b$ is the selected slot. This allows the table size to remain fixed even when the number of possible token patterns is much larger than $B$.

Hash-based compression provides an efficient way to fit memory into GPU, but it also changes the meaning of a memory entry. A physical slot no longer necessarily corresponds to a single phrase or token pattern. Instead, it may be shared by many different sequences. If these sequences are semantically related, the shared value may still provide a useful signal. However, if unrelated phrases collide in the same slot, the learned value can become an ambiguous mixture of multiple patterns. The retrieved memory signal may then be less phrase-specific, or even misleading.

This issue is particularly important for static phrase memory. Many useful knowledge units are short multi-token expressions, such as named entities, technical terms, scientific concepts, product names, organization names, and recurring domain-specific phrases. These phrases often carry meanings that are not fully captured by individual token embeddings. A phrase-level memory pathway can therefore provide a more direct representation of such semantic units. However, if multiple unrelated phrases are compressed into the same hash slot, the memory system loses the ability to preserve clean phrase-specific signals. Compared with document-level retrieval, phrase-level memory is more compact; compared with token-level lookup, it better matches semantic boundaries; but compared with hash-collided memory, it requires more precise logical entries.

Conventional Engram-style designs are therefore not yet sufficient for large-scale train-free static memory. First, memory entries are often learned from data~\cite{memorygrafting}. This means the memory table is not simply a static external database; it is part of the model optimization process. As a result, constructing or adapting memory requires additional training. Second, existing Engram memories are often model-specific. The value vectors must match the hidden space of a particular LLM, and the lookup or injection mechanism may depend on the model's architecture and hidden dimension. Third, compact GPU-resident implementations often rely on compression or hashing, which improves storage efficiency but can weaken phrase-level semantic fidelity under heavy collision.

These limitations become more serious when the goal is to build a large static memory from external corpora. A corpus-derived phrase memory may contain millions or billions of candidate entries. Such a table is too large to fit entirely in GPU memory if every phrase is represented precisely~\cite{kwon2023vllm,sheng2023flexgen,rajbhandari2021zeroinfinity}. At the same time, compressing it aggressively into a small hash table can cause unrelated phrases to share memory values. Moreover, retraining all memory entries together with the LLM would be prohibitively expensive. Static phrase knowledge is naturally suited for offline construction: once phrases are mined and encoded, many entries can remain unchanged across inference runs. This suggests that Engram should be reinterpreted not only as a neural memory module, but also as a storage-backed memory system.

This paper builds on this reinterpretation. We use Engram-style memory as the integration mechanism because it provides a compact side pathway for hidden-state injection. At the same time, we redesign its construction, storage, and prefetch assumptions. Instead of learning all memory entries through model training, TF-Engram constructs phrase memories offline. Instead of relying primarily on hash compression to fit memory into GPU, TF-Engram preserves phrase-specific logical entries and organizes them across GPU cache, host memory, and SSD. Instead of waiting for final output predictions to prefetch memory, TF-Engram uses an early-exit prediction signal from an intermediate layer to overlap memory transfer with the computation of the remaining Transformer layers.

\section{Motivation}

\subsection{Hash-Compressed Engram Weakens Static Memory Capability}

Engram-style memory is designed to provide a compact side pathway for injecting reusable knowledge into LLM hidden states. Ideally, each memory entry should represent a meaningful token pattern or phrase, and the retrieved value should provide a stable signal that helps the model predict the next token. This design is attractive because many useful knowledge units in language are phrase-level: named entities, technical terms, scientific concepts, organization names, geographic expressions, and frequent multi-token expressions. Such phrases often have stable semantics and repeatedly appear across corpora. Therefore, they are natural candidates for static memory.

However, directly storing all useful phrases is difficult. The number of possible token $n$-grams grows rapidly with corpus size, and a corpus-scale phrase table can easily contain millions or billions of candidates. If each phrase is assigned an independent memory value, the table quickly exceeds practical GPU memory limits. Existing Engram-style systems address this scalability problem by using a fixed-size hash-based table. Instead of allocating one unique slot for every possible token pattern, many token sequences are mapped into a smaller number of memory slots through hashing. This allows the memory table to remain compact and GPU-resident.

This hash-based compression solves the storage problem, but introduces a more fundamental memory-quality problem. When many phrases are mapped into a fixed-size table, different token patterns inevitably collide in the same memory slot. A single slot may therefore receive training signals from multiple unrelated phrases. The learned value in that slot no longer represents a specific phrase or semantic unit; instead, it becomes an average or mixture of all collided patterns. If the collided phrases are semantically related, the shared value may still be useful. However, if they are unrelated, the retrieved memory value becomes ambiguous. In the worst case, the memory signal can be actively misleading because the model retrieves a value influenced by a different phrase.

This issue is especially problematic for static phrase memory. The goal of static phrase memory is to preserve stable phrase-level knowledge. For example, a phrase such as \emph{New York City}, \emph{quantum field theory}, or \emph{BGP route reflector} should ideally correspond to a memory entry that preserves its specific semantic meaning. Hash collision breaks this assumption. If multiple unrelated phrases share the same slot, the table cannot provide a clean phrase-specific memory signal. Thus, the hash table may be compact, but its memory capability is weakened by semantic ambiguity.

The severity of this problem increases as the number of candidate phrases grows. Suppose $M$ distinct phrases are mapped into $B$ hash slots. When $M$ is much larger than $B$, most slots will contain multiple phrases. The expected number of occupied slots is approximately:
\begin{equation}
B \left(1 - \left(1 - \frac{1}{B}\right)^M \right),
\end{equation}
and the average number of phrases per occupied slot grows as the phrase set expands. Therefore, increasing the corpus size without increasing the table size does not linearly increase usable memory capacity. Instead, it increases collision pressure. The memory table appears to cover more phrases, but many of these phrases are no longer represented independently.

\begin{table}[t]
\centering
\caption{Hash-compressed memory creates a mismatch between represented phrase space and physical memory slots. When the candidate phrase space is much larger than the table, many phrases must share the same slot.}
\label{tab:hash_collision}
\begin{tabular}{lcc}
\toprule
Candidate Phrases & Hash Slots & Avg. Phrases per Slot \\
\midrule
1M   & 1M & 1.0 \\
100M & 1M & 100.0 \\
1B   & 1M & 1000.0 \\
\bottomrule
\end{tabular}
\end{table}

This creates a tension in existing Engram designs. On one hand, increasing the number of covered phrases is necessary for stronger memory capability. A small memory table can only capture a limited number of useful static patterns. On the other hand, if a large phrase space is compressed into a small hash table, the resulting collisions make the memory values less precise. The system therefore faces an unfavorable trade-off: either maintain a small table with limited coverage, or use hash compression to increase apparent coverage while sacrificing phrase-specific semantic clarity.

This trade-off is inconsistent with the original motivation of static memory. Static phrase knowledge should be explicit, stable, and reusable. A memory system should retrieve a value because the current phrase matches a meaningful stored entry, not because the phrase happens to collide with many other patterns in a compressed hash table. Therefore, the key limitation of existing Engram-style memory is not only that GPU memory is limited. The deeper issue is that hash-based compression uses ambiguous shared slots to approximate a much larger phrase space, which reduces the semantic fidelity of the retrieved memory.

This observation motivates a different design point. Instead of forcing a large phrase space into a small GPU-resident hash table, we should preserve phrase-specific memory entries and move the capacity burden into a storage hierarchy. In other words, Engram memory should scale by externalizing storage rather than by increasing hash collision. This allows the memory system to maintain cleaner phrase-level semantics while supporting much larger memory capacity.

\subsection{Externalizing Precise Engram Memory Introduces New Challenges}

Externalizing Engram memory is a natural way to avoid excessive hash compression. If memory entries can be stored outside GPU memory, the system no longer needs to map a massive phrase space into a small number of slots. Instead, each important phrase can have a more precise memory representation, and the table can scale with corpus size. However, this design changes the nature of Engram from a compact neural table into a storage-backed external memory system. Once the memory is externalized, three challenges must be addressed: how to construct phrase-specific memory without training a huge parameter table, how to organize the memory across GPU, DRAM, and SSD, and how to hide the latency of external lookup during autoregressive decoding.

\textbf{Challenge 1: Large precise memory cannot be trained as a conventional Engram table.}
If each phrase is assigned a distinct memory entry, the table may contain millions or billions of entries. Treating such a table as ordinary trainable parameters is impractical. Training would require large-scale optimization, significant GPU memory, and repeated access to massive corpora. It would also make the memory tightly coupled with a specific backbone model, hidden dimension, and training objective. When the corpus changes or a new domain is added, the table may need additional training. This defeats the purpose of using static phrase memory as a reusable external knowledge structure.

A scalable memory construction method should therefore avoid learning every memory entry through LLM training. Since many phrase memories correspond to stable semantic units, they can be constructed offline. The system can mine candidate phrases from large corpora, encode their meanings using an external semantic encoder, and store them as static memory entries. This construction process better matches the nature of phrase-level knowledge: the memory entry is tied to the phrase itself rather than to a hash slot shared by unrelated phrases. The challenge is to build such entries in a train-free way while still making them compatible with the LLM inference pathway.

\textbf{Challenge 2: Precise memory requires storage-aware hierarchy instead of hash-only compression.}
Removing or reducing hash compression increases memory fidelity, but also increases storage demand. A phrase-specific table can be much larger than a compressed Engram table. Keeping the entire table in GPU memory is unrealistic because GPU memory is already occupied by model weights, KV cache, activations, and runtime buffers. Therefore, precise Engram memory must be organized across multiple storage tiers. Hot entries should remain close to the model, while cold entries should be placed in cheaper and larger storage such as host DRAM or NVMe SSD.

This creates a storage-management problem. GPU memory provides low latency but limited capacity. Host DRAM provides more capacity but requires CPU--GPU data movement. SSD provides much larger capacity but has much higher random-access latency. The system must decide which entries should stay in GPU cache, which entries should be staged in DRAM, and which entries should remain on SSD. It must also support efficient lookup without turning phrase matching into random blocking I/O. Therefore, scaling precise Engram memory requires a storage-aware layout, index structure, and cache policy. The system should use SSD capacity to preserve phrase-specific memory entries, while keeping frequently accessed entries in faster tiers.

\textbf{Challenge 3: External lookup latency must be hidden before it reaches the decoding critical path.}
Even if SSD-backed memory solves the capacity problem, it introduces a latency problem. Autoregressive decoding is sequential: each generated token depends on previous tokens, and each step has a strict latency budget. If the model encounters a phrase and then waits for a blocking SSD read or CPU--GPU transfer, the entire decoding pipeline stalls. This is unacceptable for LLM serving, where per-token latency directly affects throughput and user-perceived responsiveness. Therefore, precise external memory cannot rely on naive on-demand SSD lookup.

A straightforward predictive design may attempt to use the final next-token distribution produced by the original language modeling head. However, this distribution becomes available only after all Transformer layers have completed the forward pass for the current token. At that point, there is little remaining computation with which to overlap memory movement. In other words, final-output-guided prefetching can predict future memory accesses, but it is often too late to hide SSD or host-to-device transmission latency.

The system must instead predict memory accesses earlier. Decoder-only Transformers naturally provide intermediate hidden states before the final output layer. If an auxiliary early-exit head is attached to an intermediate layer near the end of the network, such as layer $L-r$, it can produce an approximate next-token distribution before the full forward pass completes. This early prediction can be used only for memory scheduling, while the final token remains produced by the original LLM head after all layers finish. The resulting prefetch request can be issued while the remaining layers $L-r+1,\ldots,L$ are still being computed.

This creates a useful latency-hiding window. During the computation of the remaining Transformer layers, the memory system can asynchronously fetch candidate phrase entries from SSD or host DRAM and stage them into faster tiers. If the predicted phrase is actually needed by the time memory injection occurs, the lookup can be served from GPU cache or host memory rather than triggering a blocking external access. The challenge is to balance three factors: placing the early-exit head early enough to expose a useful overlap window, placing it late enough to preserve prediction quality, and limiting the top-$K$ prefetch budget to avoid excessive wasted I/O.

These challenges define the design requirements for TF-Engram. First, memory construction should be train-free and semantic, so that phrase-specific entries can be built offline without learning a large hash-compressed parameter table. Second, memory storage should be hierarchical, so that the system can preserve precise phrase entries without requiring all of them to fit in GPU memory. Third, memory access should be early-predicted and asynchronous, so that SSD and CPU--GPU transfer latency can be hidden behind the computation of remaining Transformer layers rather than exposed on the token-generation critical path. Together, these requirements motivate TF-Engram as a train-free, SSD-backed, and early-exit-guided extension of Engram-style memory.

\section{TF-Engram Design}

\begin{figure*}[t]
    \centering
    \includegraphics[width=\textwidth]{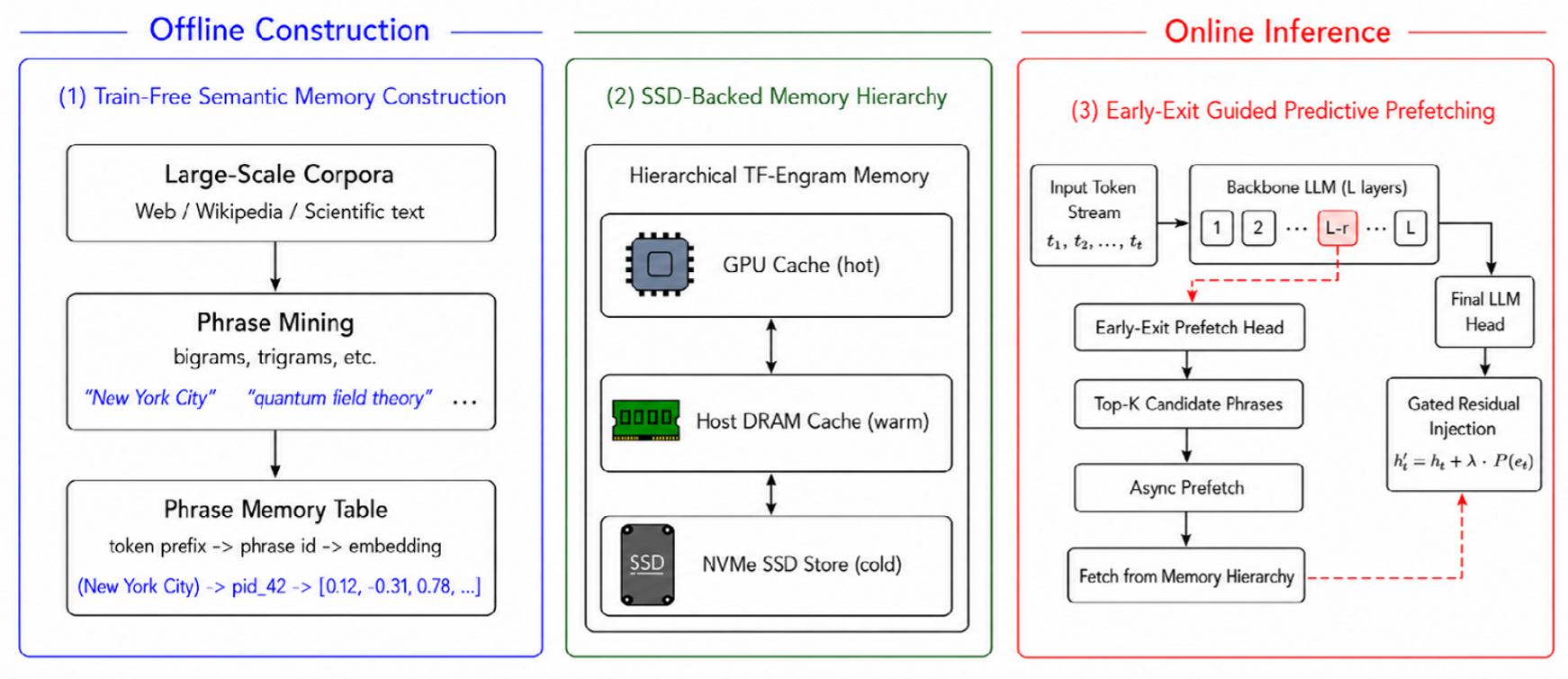}
    \caption{Technique overview of TF-Engram.}
    \label{fig:tfengram_overview}
\end{figure*}

\subsection{Overview}

Figure~\ref{fig:tfengram_overview} presents the overall design of TF-Engram. 
TF-Engram transforms Engram-style memory from a compact hash-compressed GPU table into a train-free, SSD-backed external memory system. 
Its main goal is to preserve phrase-specific static memory entries while scaling memory capacity beyond accelerator memory and hiding the latency of external memory access during autoregressive decoding.

As shown in Figure~\ref{fig:tfengram_overview}, TF-Engram consists of three components. 
The first component, \textit{Train-Free Semantic Memory Construction}, is performed offline. 
Starting from large-scale corpora such as web text, Wikipedia, and scientific text, TF-Engram mines reusable phrase candidates, such as bigrams, trigrams, and other phrase-level semantic units. 
These phrases are then organized into a phrase memory table, where each entry maps a token prefix to a phrase identifier and a phrase embedding. 
Unlike conventional Engram systems that rely on learned or hash-compressed memory slots, TF-Engram explicitly preserves phrase-specific memory entries constructed offline.

The second component, \textit{SSD-Backed Memory Hierarchy}, organizes the resulting memory table into a hierarchical storage system. 
Instead of forcing all memory entries to reside in GPU memory, TF-Engram separates memory into three tiers: a hot GPU cache, a warm host DRAM cache, and a cold NVMe SSD store. 
This hierarchy allows TF-Engram to preserve a much larger phrase memory table while keeping frequently accessed entries close to the model. 
As a result, TF-Engram scales memory capacity through hierarchical storage rather than through aggressive hash compression.

The third component, \textit{Early-Exit Guided Predictive Prefetching}, operates during online inference. 
Given an input token stream, the backbone LLM processes the current decoding step as usual. 
Before the full forward pass completes, TF-Engram taps an intermediate layer near the end of the backbone, denoted as layer $L-r$, and attaches an early-exit prefetch head. 
This head predicts top-$K$ candidate phrases that are likely to be accessed in the near future. 
TF-Engram then asynchronously prefetches the corresponding memory entries from the hierarchical memory system while the remaining Transformer layers continue computation. 
Meanwhile, the original final LLM head produces the final hidden representation, and the fetched memory is incorporated through gated residual injection. 
By overlapping memory fetch with the computation of the remaining layers, TF-Engram hides much of the latency of SSD and host-to-device transfer.

Overall, TF-Engram separates \textit{offline construction} from \textit{online inference}. 
Offline, it builds precise phrase-level memory entries without training the backbone model. 
Online, it combines hierarchical storage with early-exit-guided predictive prefetching to make large external phrase memory practical for autoregressive LLM inference.

\subsection{Train-Free Semantic Memory Construction}

The first component of TF-Engram is an offline construction pipeline for phrase-specific semantic memory. The goal is to build a large phrase memory table without training memory entries as model parameters. This differs from conventional Engram-style memory, where memory values are often learned through model training and compressed into a fixed-size hash table. TF-Engram treats phrase memory as a static knowledge structure that can be constructed from corpora, stored externally, and reused during inference.

The construction pipeline begins with phrase mining. Given a corpus $\mathcal{C}$, TF-Engram extracts candidate multi-token expressions such as bigrams, trigrams, named entities, technical terms, scientific concepts, organization names, and domain-specific phrases. Candidate phrases can be selected using frequency thresholds, statistical association, named-entity recognition, or corpus-specific filters. The objective is not to enumerate every possible token sequence, but to select phrases that correspond to reusable semantic units.

After phrase extraction, each phrase is tokenized using the target LLM tokenizer:
\begin{equation}
\tau_i = \mathrm{Tokenizer}(\phi_i).
\end{equation}
The tokenized form $\tau_i$ is used for inference-time matching, while the original phrase string $\phi_i$ is used for semantic encoding and metadata management. TF-Engram can filter out phrases with unstable tokenization, excessive length, or low corpus support. This step ensures that the phrase memory table remains useful for token-stream lookup during LLM inference.

TF-Engram then encodes each phrase using a frozen semantic encoder, instantiated with Qwen3-Embedding in our implementation~\cite{zhang2025qwen3embedding}:
\begin{equation}
e_i = E(\phi_i),
\end{equation}
where $E(\cdot)$ is an external embedding model and $e_i$ is the semantic memory vector. This is the central difference from hash-compressed Engram. Instead of assigning many unrelated phrases to a shared memory slot and learning a mixed value, TF-Engram assigns each selected phrase its own semantic representation. As a result, the memory value remains tied to the phrase itself rather than to an ambiguous hash bucket.

To support efficient inference-time lookup, TF-Engram builds a two-level index:
\begin{equation}
\mathrm{token\ prefix} \rightarrow \mathrm{phrase\ id} \rightarrow \mathrm{memory\ address}.
\end{equation}
The first level maps token prefixes or suffix candidates to phrase identifiers. During inference, the phrase matcher only needs to verify a compact candidate list instead of scanning the full memory table. The second level maps phrase identifiers to memory metadata, including phrase length, semantic vector address, frequency score, storage tier, and physical offset.

When multiple phrases match the current context, TF-Engram uses a deterministic selection policy. A simple policy is longest-match selection, where longer phrases are preferred because they usually correspond to more complete semantic units. If several phrases have the same length, TF-Engram can use frequency, semantic confidence, or corpus priority as tie-breakers. If multiple matched phrases are retained, their memory vectors can be aggregated:
\begin{equation}
e_{\phi} = \sum_{i \in \mathcal{P}_t} \alpha_i e_i,
\end{equation}
where $\mathcal{P}_t$ is the set of matched phrases at position $t$ and $\alpha_i$ is determined by phrase length, frequency, or confidence.

This construction is train-free in the sense that the phrase memory entries are not optimized through LLM training. The memory table is built by corpus processing, tokenization, and frozen semantic encoding. Adding a new corpus only requires mining and encoding new phrases, rather than retraining the backbone model or re-optimizing a large memory parameter table.

\subsection{SSD-Backed Memory Hierarchy}

The second component of TF-Engram is an SSD-backed memory hierarchy. The purpose of this hierarchy is to preserve phrase-specific memory entries without requiring the entire table to fit in GPU memory. Instead of using hash compression to force a large phrase space into a small GPU-resident table, TF-Engram keeps precise logical entries and uses storage hierarchy to absorb the capacity demand.

TF-Engram organizes memory into three tiers:
\begin{equation}
\mathcal{M} = \mathcal{M}*{gpu} \cup \mathcal{M}*{dram} \cup \mathcal{M}*{ssd},
\end{equation}
where $\mathcal{M}*{gpu}$ stores hot entries, $\mathcal{M}*{dram}$ stores warm entries, and $\mathcal{M}*{ssd}$ stores cold entries. The GPU tier provides the lowest latency and is reserved for the most frequently accessed entries. The DRAM tier provides a larger staging cache and metadata store. The SSD tier stores the full memory table and absorbs the long tail of phrase entries.

A memory lookup proceeds through the hierarchy. When a phrase is matched, TF-Engram first checks whether the corresponding memory value is already in GPU cache. If so, the value can be used immediately for hidden-state injection. If the entry is in host DRAM, it is staged to GPU memory. If the entry is only on SSD, the system issues an asynchronous read and inserts the entry into a faster tier when it arrives. To reduce random I/O overhead, memory entries can be stored in pages or blocks, allowing multiple nearby vectors and metadata records to be fetched together.

The cache policy is important because Engram memory access is sparse and data-dependent. Unlike parameter offloading, where access order is largely determined by model layer order, phrase memory access depends on the token stream and generated content. TF-Engram can rank memory entries using access frequency, recency, phrase length, domain priority, and prefetch confidence. Frequently used general phrases remain close to the model, while rare long-tail phrases stay on SSD until needed.

The SSD layout is designed to reduce exposed storage latency. Memory vectors are stored in contiguous arrays, and phrase metadata maps phrase identifiers to SSD offsets. Phrases that are likely to be accessed together can be placed nearby, such as phrases sharing a token prefix or belonging to the same corpus domain. This improves spatial locality for batched prefetch requests. TF-Engram can also use memory-mapped files or asynchronous I/O interfaces to overlap SSD reads with LLM computation.

This hierarchy allows TF-Engram to scale memory capacity independently from GPU memory. GPU memory is reserved for model weights, KV cache, activations, runtime buffers, and the hottest phrase entries. SSD storage holds the large static phrase memory. Therefore, TF-Engram can preserve phrase-specific entries without collapsing them into ambiguous hash-collision slots.

\subsection{Early-Exit Guided Predictive Prefetching}
\label{sec:early_exit_prefetch}

The third component of TF-Engram is \emph{Early-Exit Guided Predictive Prefetching}. SSD-backed memory provides large memory capacity, but a blocking access to host DRAM or NVMe SSD during autoregressive decoding can directly stall token generation. Therefore, TF-Engram must not only store a large static memory table, but also move likely useful phrase entries into faster memory tiers before they are accessed.

\begin{figure}[t]
    \centering
    \includegraphics[width=\linewidth]{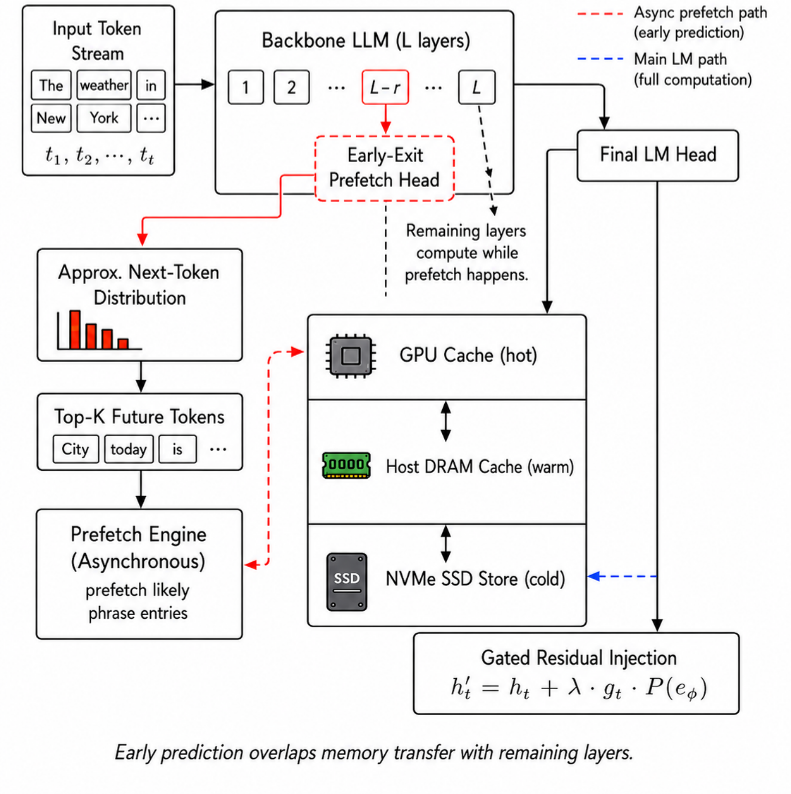}
    \caption{Early-exit guided predictive prefetching.}
    \Description{An overview of early-exit guided predictive prefetching. An intermediate Transformer layer predicts likely future tokens, which are converted into candidate phrase entries and prefetched asynchronously from SSD or host DRAM into GPU cache before gated residual injection.}
    \label{fig:early_exit_prefetch}
\end{figure}

Figure~\ref{fig:early_exit_prefetch} shows the overall design. A straightforward design would use the final next-token distribution produced by the original LM head to decide which entries to prefetch. However, this distribution is only available after all Transformer layers complete. At that point, there is little remaining computation with which to overlap the memory transfer. TF-Engram instead predicts future memory accesses from an intermediate layer. The early prediction is used only for memory scheduling, while the final generated token is still produced by the original LM head after the full forward pass.

For an LLM with $L$ Transformer layers, TF-Engram places an auxiliary early-exit prefetch head at layer $L-r$, where $r$ is the number of remaining layers. Given the intermediate hidden state $h_t^{L-r}$ at decoding step $t$, the early-exit head produces an approximate next-token distribution:
\begin{equation}
p_{\mathrm{early}}(x_{t+1} \mid x_{\leq t})
=
\mathrm{softmax}
\left(
W_{\mathrm{pref}} h_t^{L-r}
\right).
\end{equation}
Here, $W_{\mathrm{pref}}$ denotes the projection used by the prefetch head. The resulting distribution is not used for generation. It only predicts likely future memory accesses. The final output token is still computed by the original LM head after all $L$ Transformer layers finish.

TF-Engram then selects the top-$K$ likely future tokens from the early distribution:
\begin{equation}
\mathcal{T}_t^{\mathrm{early}}
=
\mathrm{TopK}
\left(
p_{\mathrm{early}}(x_{t+1} \mid x_{\leq t})
\right).
\end{equation}
These candidate tokens are combined with the recent token history to infer possible future phrases. Using the token-to-phrase index, TF-Engram maps the predicted token candidates to a set of candidate phrase entries:
\begin{equation}
\mathcal{Q}_t
=
\left\{
\phi_i
\mid
\tau_i \in \mathcal{M}(\mathcal{T}_t^{\mathrm{early}})
\right\}.
\end{equation}
Here, $\mathcal{M}(\cdot)$ maps the early predicted token candidates to phrase entries through the token-to-phrase index.
The main benefit of early-exit prediction is latency overlap. Once the prefetch request is issued at layer $L-r$, the backbone LLM continues computing the remaining layers $L-r+1,\ldots,L$. During this time, TF-Engram can move candidate entries from NVMe SSD to host DRAM, or from host DRAM to the GPU cache. The available latency-hiding window can be approximated as:
\begin{equation}
T_{\mathrm{hide}}
\approx
T_{\mathrm{compute}}
\left(
L-r+1,\ldots,L
\right),
\end{equation}
where $T_{\mathrm{compute}}(L-r+1,\ldots,L)$ denotes the computation time of the remaining Transformer layers. If a prefetched entry is later selected by the actual phrase lookup, the access can be served from a faster tier instead of triggering a blocking SSD access.

The early-exit prefetch head can be implemented in either a train-free or lightly calibrated form. In the strict train-free setting, TF-Engram reuses a frozen projection derived from the backbone model to estimate the early next-token distribution, without updating the backbone model or the Engram table. If a small calibration stage is allowed, $W_{\mathrm{pref}}$ can be trained on unlabeled text using the standard next-token objective while keeping both the backbone model and the memory table frozen. In both cases, the early-exit head affects only prefetch scheduling and does not change the final token probabilities produced by the LLM.

The position of the early-exit head introduces a trade-off between prediction quality and overlap opportunity. Placing the head earlier provides a larger latency-hiding window, but the intermediate hidden state may produce a less accurate next-token prediction. Placing the head later improves prediction quality, but leaves less remaining computation to hide memory-transfer latency. TF-Engram therefore treats the early-exit layer position, the top-$K$ prefetch budget, and the prefetch queue size as tunable system parameters.

TF-Engram further prioritizes prefetch candidates according to both prediction likelihood and expected memory benefit. Each candidate phrase $\phi_i$ is assigned a priority score:
\begin{equation}
s_i
=
p_i \cdot r_i \cdot c_i,
\end{equation}
where $p_i$ represents the probability of the predicted token path, $r_i$ estimates the usefulness of the phrase entry, and $c_i$ reflects the expected benefit of prefetching the entry from its current storage tier. For example, an entry currently residing on SSD receives a higher prefetch benefit than an entry already cached in GPU memory. High-priority candidates are prefetched first, while low-priority candidates are dropped when the I/O budget or queue capacity is exhausted.

Overall, early-exit guided predictive prefetching enables TF-Engram to use large SSD-backed memory without exposing the full SSD access latency to decoding. By predicting likely future phrase accesses before the forward pass completes, TF-Engram overlaps memory transfer with the remaining Transformer computation and makes large static Engram tables practical for low-latency autoregressive inference.

\section{Evaluation}

We evaluate TF-Engram from four perspectives. First, we study end-to-end effectiveness by comparing TF-Engram with the original Qwen3-0.6B backbone and a parameter-matched LoRA baseline on downstream benchmarks. This evaluates whether external phrase-level memory can improve model quality without updating the backbone parameters. Second, we analyze the overhead introduced by TF-Engram, including offline construction cost, storage footprint, runtime memory usage, throughput, and decoding latency. Third, we study the sensitivity of TF-Engram to three key parameters: memory injection scale, early-exit layer position, and top-$K$ prefetch budget. Finally, we conduct ablation studies to isolate the contribution of train-free semantic memory, SSD-backed hierarchy, and early-exit guided prefetching.

\subsection{Experimental Setup}

We evaluate TF-Engram on a single-node workstation. Table~\ref{tab:hardware_setup} summarizes the hardware configuration. The machine contains a high-end desktop CPU, a single NVIDIA RTX 5090 GPU, host DRAM, and a local NVMe SSD. This setup allows us to evaluate whether TF-Engram can support large external phrase memory under limited accelerator memory, without relying on a multi-GPU server.

\begin{table}[t]
\centering
\caption{Hardware configuration used in our evaluation.}
\label{tab:hardware_setup}
\begin{tabular}{ll}
\toprule
Component & Configuration \tabularnewline
\midrule
CPU & Intel Core i9-14900K \tabularnewline
CPU cores / threads & 24 cores / 32 threads \tabularnewline
Max CPU frequency & 6.0 GHz \tabularnewline
GPU & NVIDIA GeForce RTX 5090 \tabularnewline
GPU memory & 32GB \tabularnewline
Host memory & 64GB \tabularnewline
SSD storage & 516GB \tabularnewline
NUMA nodes & 1 \tabularnewline
\bottomrule
\end{tabular}
\end{table}

Unless otherwise stated, we use Qwen3-0.6B\cite{yang2025qwen3} as the backbone model. The original Qwen3-0.6B without external memory is used as the backbone baseline. For end-to-end model-quality comparison, we additionally include a parameter-matched LoRA baseline. The LoRA baseline represents a lightweight trainable adaptation method with a comparable additional trainable-parameter budget, while TF-Engram keeps the backbone model frozen and introduces an external phrase-level memory pathway.

TF-Engram constructs phrase memory offline from a mixture of FineWeb-Edu, Wikipedia, C4 English, arXiv, and PubMed. These corpora cover general web text, encyclopedic factual knowledge, commonsense text, scientific terminology, and biomedical terminology. The constructed phrase entries are stored in a GPU--DRAM--SSD hierarchy: GPU memory caches hot entries, host DRAM caches warm entries, and NVMe SSD stores the full phrase table. The memory size, prefetch budget, and early-exit layer position are specified in the corresponding experiments.

For end-to-end effectiveness, we evaluate downstream accuracy on MMLU, ARC-Challenge, HellaSwag, PIQA, WinoGrande, BoolQ, OpenBookQA, SciQ, TruthfulQA-MC2, and LAMBADA. For overhead analysis, we report offline construction time, storage footprint, GPU memory usage, throughput, and per-token latency. For sensitivity analysis, we vary memory injection scale, early-exit layer position, and top-$K$ prefetch budget. For ablation studies, we compare TF-Engram variants that remove or replace individual components, including SSD-backed memory without prefetching, final-head-guided prefetching, and the full early-exit-guided design.

\subsection{End-to-End Effectiveness}

We first evaluate whether TF-Engram improves end-to-end model quality on representative knowledge, reasoning, and language understanding benchmarks. We compare three settings: the original Qwen3-0.6B backbone, a parameter-matched LoRA baseline, and TF-Engram. The LoRA baseline uses a comparable trainable-parameter budget and represents a lightweight trainable adaptation method. In contrast, TF-Engram keeps the backbone model frozen and introduces external phrase-level memory. Table~\ref{tab:e2e_main} reports the detailed results.

\begin{table}[t]
\centering
\small
\caption{End-to-end performance comparison across downstream benchmarks.}
\label{tab:e2e_main}
\begin{tabular}{lccc}
\toprule
Benchmark & Qwen3-0.6B & Parameter-Matched LoRA & TF-Engram \tabularnewline
\midrule
MMLU & 53.7 & 55.3 & 56.2 \tabularnewline
ARC-Challenge & 42.0 & 44.1 & 45.3 \tabularnewline
HellaSwag & 51.0 & 52.9 & 50.6 \tabularnewline
PIQA & 70.0 & 69.5 & 72.1 \tabularnewline
WinoGrande & 63.0 & 64.3 & 62.8 \tabularnewline
BoolQ & 76.0 & 75.6 & 78.4 \tabularnewline
OpenBookQA & 32.0 & 34.3 & 35.1 \tabularnewline
SciQ & 95.0 & 94.8 & 95.4 \tabularnewline
TruthfulQA-MC2 & 45.0 & 46.5 & 48.2 \tabularnewline
LAMBADA & 48.0 & 49.6 & 50.3 \tabularnewline
\midrule
Average & 57.6 & 58.7 & 59.4 \tabularnewline
\bottomrule
\end{tabular}
\end{table}

Overall, TF-Engram achieves the best average performance among the three methods. The average score improves from 57.6 for the original Qwen3-0.6B backbone to 58.7 for the parameter-matched LoRA baseline and further to 59.4 for TF-Engram. This corresponds to an absolute improvement of 1.8 points over the backbone and 0.7 points over LoRA.

TF-Engram improves over the original backbone on 8 out of 10 benchmarks. The largest gains are observed on ARC-Challenge, TruthfulQA-MC2, OpenBookQA, MMLU, BoolQ, and LAMBADA. These results suggest that phrase-level external memory is especially useful for knowledge-intensive and factual evaluation tasks. TF-Engram also improves PIQA and SciQ, indicating that the retrieved memory signal can benefit tasks beyond factual recall alone.

Compared with the parameter-matched LoRA baseline, TF-Engram achieves better results on most benchmarks, including MMLU, ARC-Challenge, PIQA, BoolQ, OpenBookQA, SciQ, TruthfulQA-MC2, and LAMBADA. This result indicates that external phrase-level memory can provide a complementary and effective alternative to lightweight trainable adaptation. In particular, the gains on MMLU, ARC-Challenge, and TruthfulQA-MC2 suggest that TF-Engram is better aligned with knowledge-intensive tasks where reusable phrase-level memory can provide useful semantic signals.

TF-Engram underperforms LoRA on HellaSwag and WinoGrande, and slightly underperforms the original backbone on these two benchmarks. This suggests that static phrase memory does not benefit all task types equally. Tasks that rely more heavily on contextual commonsense completion or sentence-level disambiguation may be less aligned with phrase-level memory retrieval. Nevertheless, these drops are limited and do not offset the overall average improvement.

These results show that TF-Engram improves end-to-end performance while keeping the backbone model frozen. Compared with parameter-matched LoRA, TF-Engram achieves higher average accuracy by using external phrase-level memory instead of directly updating model weights.

\subsection{Overhead Analysis}
\label{sec:overhead}

We further evaluate the cost of TF-Engram from three perspectives: offline construction cost, runtime resource overhead, and decode latency overhead.

\begin{figure}[t]
\centering
\includegraphics[width=\linewidth]{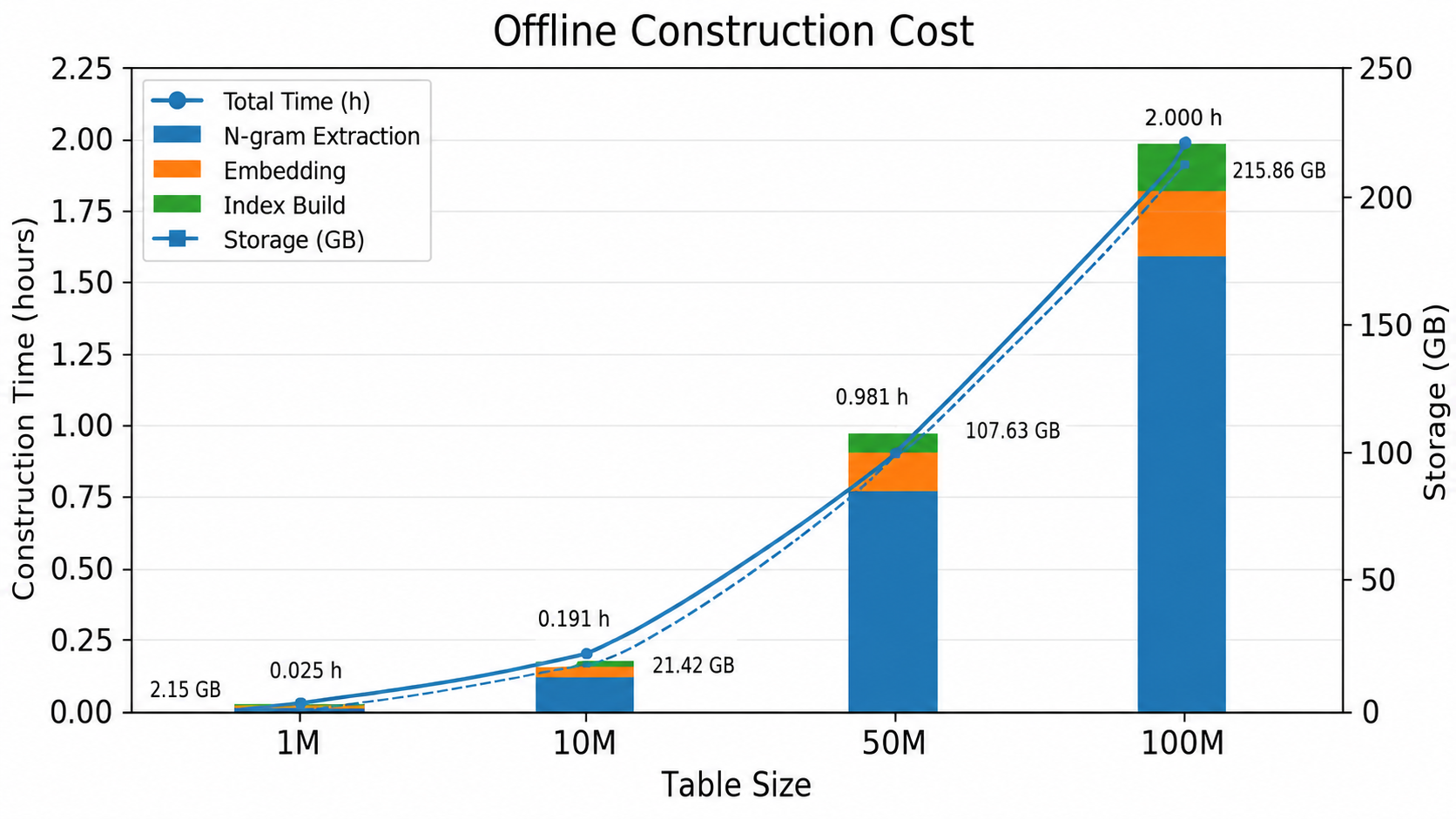}
\caption{Offline construction cost of TF-Engram.}
\label{fig:offline_construction_cost}
\end{figure}

\textbf{Offline construction cost.}
Figure~\ref{fig:offline_construction_cost} shows the offline construction cost when building TF-Engram tables with 1M to 100M entries. The construction process is dominated by tokenizer-level n-gram extraction, while embedding generation and index construction contribute only a small fraction of the total time. For a 100M-entry table, the total construction time is approximately 2.0 hours, among which n-gram extraction accounts for 1.612 hours. Embedding generation and index construction take only 0.231 hours and 0.157 hours, respectively. This indicates that TF-Engram can be constructed with moderate offline cost, and that the main bottleneck is corpus-level n-gram processing rather than vector generation or index building. The storage cost scales nearly linearly with table size, reaching 215.86 GB for a 100M-entry table with 1024-dimensional FP16 vectors.

\begin{figure}[t]
\centering
\includegraphics[width=\linewidth]{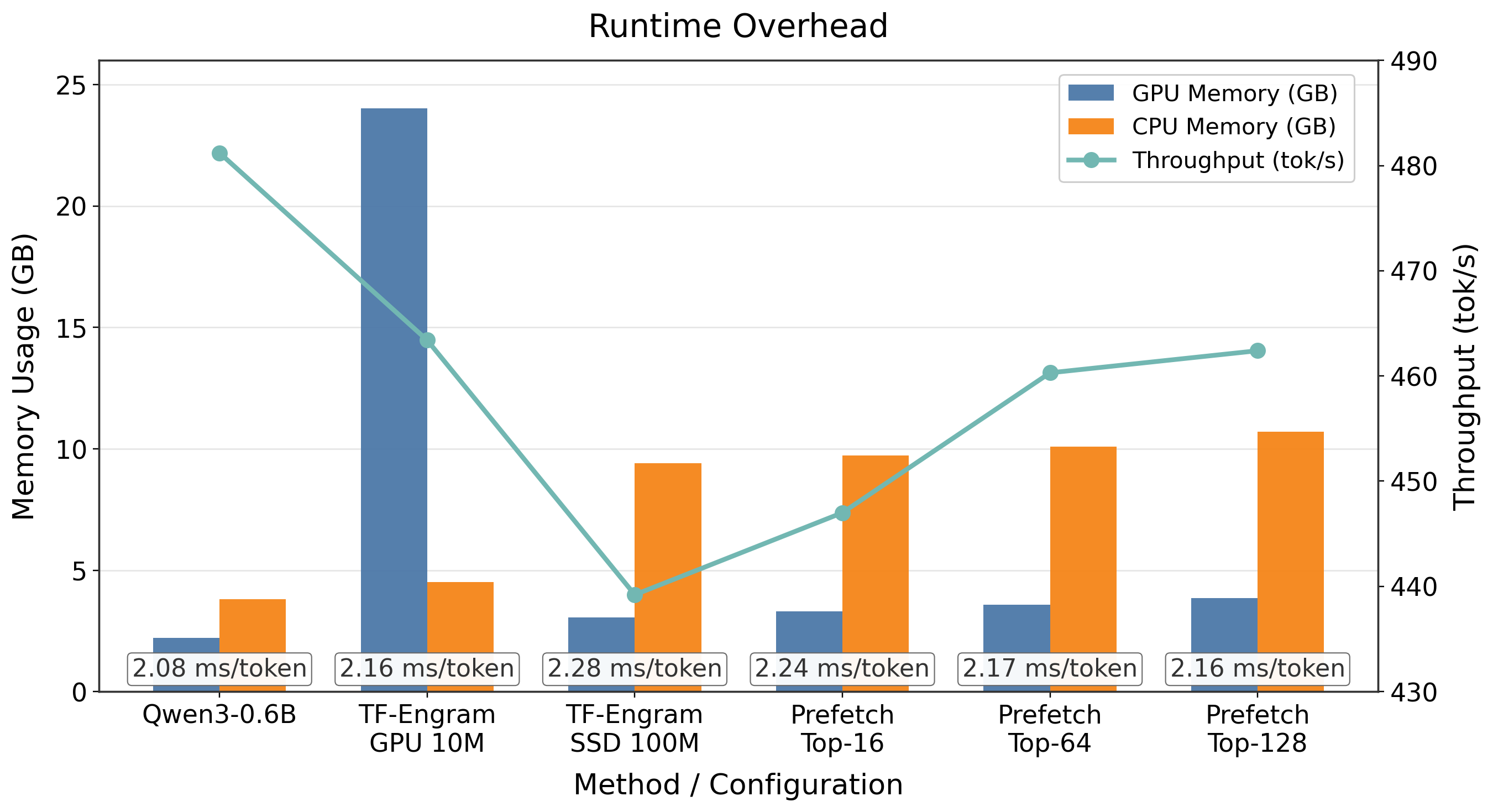}
\caption{Runtime resource overhead of TF-Engram.}
\label{fig:runtime_overhead}
\end{figure}

\textbf{Runtime resource overhead.}
Figure~\ref{fig:runtime_overhead} reports the runtime memory usage and throughput of different TF-Engram configurations. A GPU-resident 10M-entry table incurs a large GPU memory footprint, increasing GPU memory usage to 24.03 GB. In contrast, the SSD-backed 100M-entry table reduces GPU memory usage to 3.06 GB, while shifting the storage burden to CPU memory and SSD-backed access. This demonstrates the main benefit of the SSD-backed design: TF-Engram can support a much larger memory table without keeping the full table in GPU memory.

The SSD-backed configuration without prefetching reduces throughput from 481.2 tokens/s to 439.2 tokens/s. With prediction-guided prefetching, throughput is gradually recovered. Prefetching with Top-64 improves throughput to 460.3 tokens/s, and Top-128 further improves it to 462.4 tokens/s. This shows that prefetching effectively hides part of the SSD access overhead, while the remaining throughput gap mainly comes from memory lookup and data movement during decoding.

\begin{figure}[t]
\centering
\includegraphics[width=\linewidth]{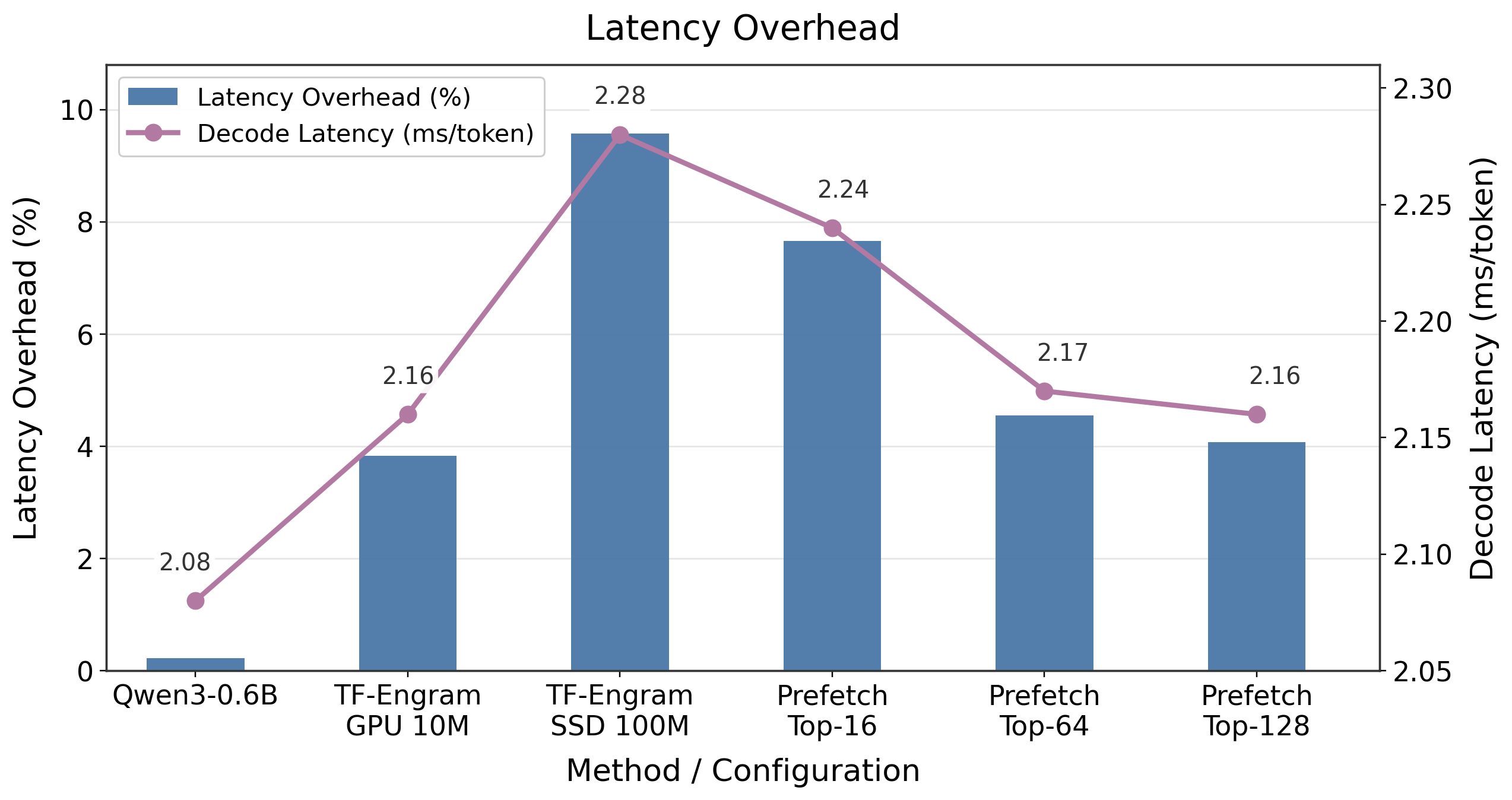}
\caption{Decode latency overhead of TF-Engram.}
\label{fig:latency_overhead}
\end{figure}

\textbf{Decode latency overhead.}
Figure~\ref{fig:latency_overhead} further breaks down the decode latency overhead. The baseline Qwen3-0.6B model achieves 2.08 ms/token. A GPU-resident 10M-entry TF-Engram table increases the latency to 2.16 ms/token, corresponding to a 3.83

Prediction-guided prefetching substantially reduces this overhead. With Top-16 prefetching, the latency decreases to 2.24 ms/token. Increasing the prefetch budget to Top-64 further reduces the latency to 2.17 ms/token, and Top-128 achieves 2.16 ms/token. The latency overhead is therefore reduced from 9.57

\subsection{Sensitivity Analysis}
\label{sec:sensitivity}

\begin{figure}[t]
\centering
\includegraphics[width=\linewidth]{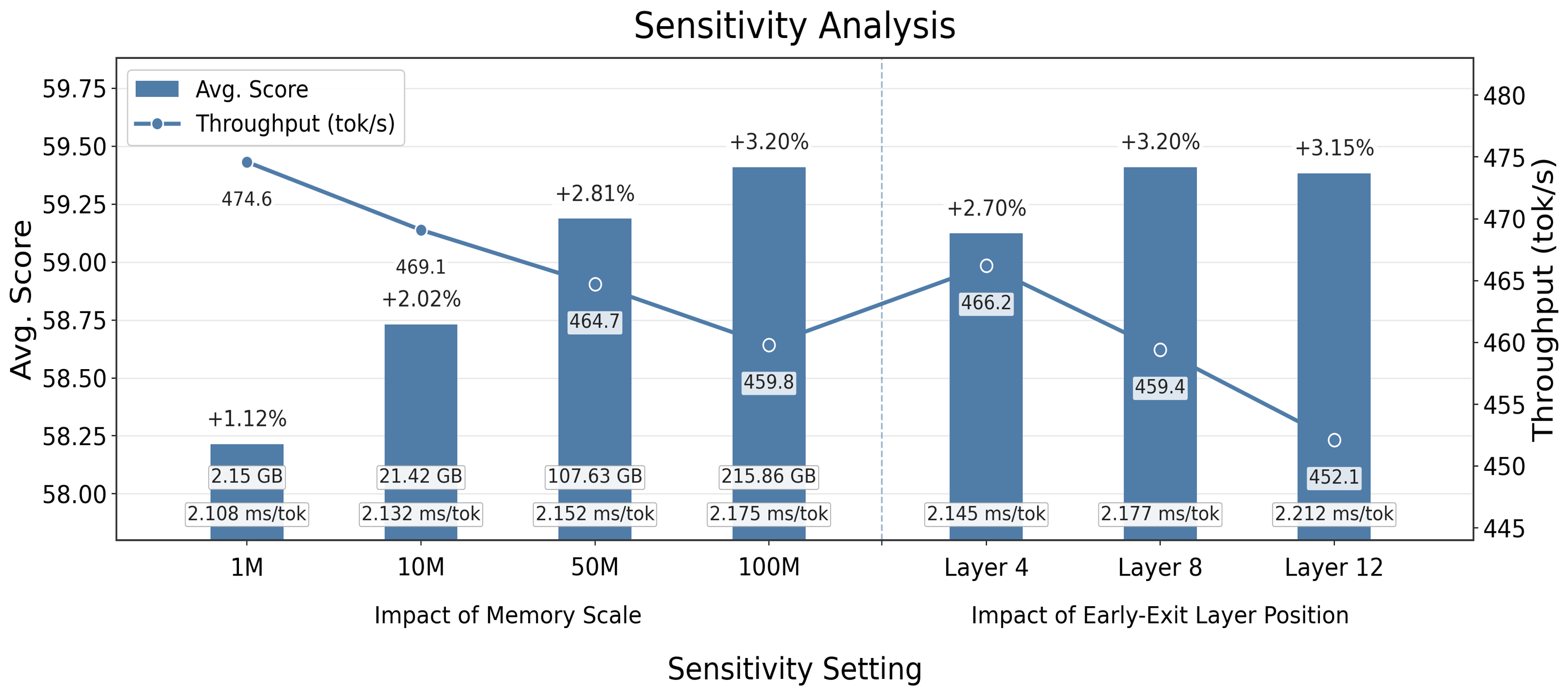}
\caption{Sensitivity analysis of TF-Engram.}
\label{fig:sensitivity_analysis}
\end{figure}

We further conduct a sensitivity analysis to understand how TF-Engram behaves under different memory scales and early-exit layer positions. The results are shown in Figure~\ref{fig:sensitivity_analysis}.

\textbf{Impact of memory scale.}
We first vary the memory table size from 1M to 100M entries. As the memory scale increases, the average score improves consistently from 58.2145 to 59.4092, corresponding to an improvement from 1.12

The benefit, however, comes with moderate runtime overhead. Decode latency increases from 2.108 ms/token at 1M entries to 2.175 ms/token at 100M entries, while throughput decreases from 474.6 tokens/s to 459.8 tokens/s. This trend reflects the expected trade-off between memory coverage and runtime cost. Even at 100M entries, the throughput degradation remains moderate, suggesting that TF-Engram can scale to large memory tables while maintaining practical decoding efficiency.

\textbf{Impact of early-exit layer position.}
We then evaluate different early-exit layer positions while fixing the memory scale to 100M entries and the prefetch budget to Top-64. The best result is obtained at Layer 8, achieving an average score of 59.4092 and a 3.20

This shows that the early-exit layer position affects both prediction quality and runtime efficiency. A shallow exit layer has lower latency but provides weaker predictive signals for prefetching. A deeper exit layer may produce more accurate predictions, but it introduces additional computation and reduces throughput. Layer 8 provides the best balance between prediction quality and overhead, achieving the highest average score while keeping decode latency at 2.177 ms/token.

Overall, the sensitivity analysis shows that TF-Engram benefits from larger memory scales and requires a moderate early-exit layer for effective prefetching. The 100M-entry table with Layer 8 early exit provides the best accuracy-efficiency trade-off in our setting.

\section{Related Work}
\label{sec:related_work}

\textbf{Retrieval- and memory-augmented language models.}
Prior work augments language models with external knowledge through retrieval or non-parametric memory, such as RAG, REALM, RETRO, Atlas, and kNN-LM~\cite{lewis2020rag, guu2020realm, borgeaud2022retro, izacard2022atlas, khandelwal2020knnlm}. These methods show that external corpora or datastores can improve language modeling and knowledge-intensive tasks. TF-Engram differs from them by constructing a static phrase-level memory table and injecting retrieved entries into hidden states, rather than appending retrieved text to the prompt or interpolating output probabilities.

\textbf{Parameter-efficient and train-free adaptation.}
Parameter-efficient methods such as adapters, prefix tuning, prompt tuning, and LoRA adapt pretrained models with a small number of trainable parameters~\cite{houlsby2019adapters, li2021prefixtuning, lester2021prompttuning, hu2021lora}. In contrast, TF-Engram targets a train-free setting where the backbone model remains frozen and the memory table is built offline from external corpora. This shifts adaptation from gradient-based parameter updates to inference-time access to static external memory.

\textbf{LLM serving, offloading, and prefetching.}
Recent LLM serving systems improve inference efficiency through batching, scheduling, KV-cache management, and memory offloading~\cite{yu2022orca, kwon2023vllm, sheng2023flexgen}. TF-Engram addresses a complementary problem: serving a large static semantic memory table during autoregressive decoding. Since SSD or host-memory access can stall token generation, TF-Engram introduces an SSD--DRAM--GPU memory hierarchy and uses early-exit guided predictive prefetching to overlap memory transfer with the remaining Transformer computation.

\textbf{Vector indexing and memory tables.}
Large embedding stores are commonly supported by vector indexing and approximate nearest-neighbor search techniques, such as product quantization and graph-based indexes~\cite{jegou2011pq, johnson2019faiss, malkov2018hnsw}. TF-Engram is related to these systems, but its workload is different: memory lookup is tightly coupled with token-by-token LLM inference, and retrieved entries are used for hidden-state injection rather than document retrieval. Therefore, TF-Engram jointly considers memory construction, storage hierarchy, runtime lookup, and prefetching for low-latency LLM inference.

\section{Conclusion}
\label{sec:conclusion}

This paper presents TF-Engram, a train-free static memory system for large language models. TF-Engram constructs phrase-level memory from external corpora and injects retrieved memory entries into the hidden states of a frozen backbone model, avoiding model-parameter updates or expensive fine-tuning. To make this memory scalable and efficient, TF-Engram combines an SSD-backed GPU--DRAM--SSD memory hierarchy with early-exit guided predictive prefetching, allowing large phrase tables to be stored beyond GPU memory while hiding much of the external-memory access latency.

Our evaluation shows that TF-Engram improves downstream performance while introducing moderate system overhead. A 100M-entry table can be constructed in approximately two hours, with tokenizer-level n-gram extraction dominating the offline cost. At runtime, SSD-backed TF-Engram substantially reduces GPU memory usage compared with GPU-resident memory, and predictive prefetching recovers much of the throughput loss while keeping decode latency overhead modest. Sensitivity analysis further shows that larger memory scales improve average task performance, and that a moderate early-exit layer provides the best trade-off between prediction quality and latency hiding.

Overall, TF-Engram demonstrates that static external memory can be integrated into LLM inference as a scalable, train-free, and low-overhead system component. By decoupling knowledge storage from dense model parameters, TF-Engram provides a practical path toward scalable memory augmentation for large language models.

\bibliographystyle{ACM-Reference-Format}
\bibliography{sample}

\end{document}